\documentclass[conference]{IEEEtran}
\IEEEoverridecommandlockouts
\usepackage{cite}
\usepackage{amsmath,amssymb,amsfonts}
\usepackage{algorithm}
\usepackage{algpseudocode}
\usepackage{graphicx}
\usepackage{textcomp}
\usepackage{xcolor}
\usepackage{multirow}
\usepackage{subfig}
\usepackage{url}

\def\BibTeX{{\rm B\kern-.05em{\sc i\kern-.025em b}\kern-.08em
    T\kern-.1667em\lower.7ex\hbox{E}\kern-.125emX}}
\begin{document}

\makeatletter
\newcommand{\linebreakand}{%
  \end{@IEEEauthorhalign}
  \hfill\mbox{}\par
  \mbox{}\hfill\begin{@IEEEauthorhalign}
}
\makeatother

\title{PaCKD: \underline{Pa}ttern-\underline{C}lustered
\underline{K}nowledge \underline{D}istillation for Compressing Memory Access Prediction Models
}
\author{
    \IEEEauthorblockN{Neelesh Gupta\IEEEauthorrefmark{2}*,
                      Pengmiao Zhang\IEEEauthorrefmark{2}*,
                      Rajgopal Kannan\IEEEauthorrefmark{3},
                      Viktor Prasanna\IEEEauthorrefmark{2}}
    \IEEEauthorblockA{\IEEEauthorrefmark{2}University of Southern California, \IEEEauthorrefmark{3}DEVCOM Army Research Lab}
    \IEEEauthorrefmark{2}\{neeleshg, pengmiao, prasanna\}@usc.edu, \IEEEauthorrefmark{3}rajgopal.kannan.civ@army.mil
    \thanks{Distribution Statement A: Approved for public release. Distribution is unlimited.}
}

\newcommand{\ourwork}{PaCKD}

\maketitle
\begingroup\renewcommand\thefootnote{* }
\footnotetext{These authors contributed equally.}
\endgroup

\begin{abstract}
Deep neural networks (DNNs) have proven to be effective models for accurate Memory Access Prediction (MAP), a critical task in mitigating memory latency through data prefetching.
However, existing DNN-based MAP models suffer from the challenges such as significant physical storage space and poor inference latency, primarily due to their large number of parameters. 
These limitations render them impractical for deployment in real-world scenarios.
In this paper, we propose \ourwork, a Pattern-Clustered Knowledge Distillation approach to compress MAP models while maintaining the prediction performance.
The \ourwork~approach encompasses three steps: clustering memory access sequences into distinct partitions involving similar patterns, training large pattern-specific teacher models for memory access prediction for each partition, and training a single lightweight student model by distilling the knowledge from the trained pattern-specific teachers. 
We evaluate our approach on LSTM, MLP-Mixer, and ResNet models, as they exhibit diverse structures and are widely used for image classification tasks in order to test their effectiveness in four widely used graph applications.
Compared to the teacher models with 5.406M parameters and an F1-score of 0.4626, our student models achieve a 552$\times$ model size compression while maintaining an F1-score of 0.4538 (with a 1.92\% performance drop).
Our approach yields an 8.70\% higher result compared to student models trained with standard knowledge distillation and an 8.88\% higher result compared to student models trained without any form of knowledge distillation.
\end{abstract}


\section{Introduction}

Memory access prediction (MAP) models play a crucial role in modern computer systems, addressing the need to optimize memory operations and overcome performance limitations imposed by the "memory wall". The memory wall refers to the widening disparity between processor speeds and memory access times leading to significant latency issues and hampering overall system performance~\cite{wulf1995hitting, carvalho2002gap}. To combat these challenges, MAP models based on deep neural networks (DNNs) are designed to prefetch data in advance, effectively reducing latency and enhancing system efficiency by proactively loading data into the cache~\cite{baer1991cache}. This proactive data prefetching strategy improves the instructions per cycle (IPC) by anticipating future data requests and fetching the data before it is actually needed~\cite{wiel1997prefetch}.

Existing MAP models typically exhibit substantial memory system requirements and encounter significant inference latency issues.
These limitations pose challenges when deploying these models in memory-constrained systems or applications that prioritize low inference latency~\cite{choi2021survey}.
Additionally, existing MAP models may not effectively handle the dynamic and diverse memory access patterns encountered in complex workloads, such as those found in graph applications, resulting in sub-optimal prefetching performance~\cite{hashemi2018mempatterns}.
Prior work has explored the compression of LSTM prefetchers, but despite achieving some level of compression, the resulting models still remain relatively large and focus on enabling online training~\cite{srivastava2019predicting}. 
In contrast, our approach aims to address these limitations by targeting offline training with a highly compressed model that undergoes extensive training to ensure performance and efficiency.
By leveraging the benefits of offline training and achieving significant compression, we aim to enhance the practicality and performance of MAP models for various target platforms.

Compressing MAP models while preserving performance poses significant challenges. 
First, decreasing the size of a neural network model can lead to a notable drop in accuracy, thus impacting its ability to accurately predict memory access patterns~\cite{zeng2017lstm}. 
Since few layers imply quick weight saturation, smaller models fail to accurately capture complex relationships between the input and output. 
Compressed neural network models face the problem of inattention or carelessness where they may only work well for commonly appearing patterns and falter for infrequent patterns~\cite{hooker2019compressed}. 
This problem is further exacerbated in memory access prediction, where the intricate patterns involved pose additional challenges for smaller machine learning models to adequately capture and generalize the underlying patterns.
Second, the memory access patterns vary across different execution stages of an application, hindering model training and overall knowledge learnt through features. For example, graph processing applications with multiple phases~\cite{lakhotia2020gpop,xstream,pcpm} make it difficult to train a general ML model that performs well across all phases~\cite{french1999catastrophic, kirkpatrick2017overcoming}. 

We propose \ourwork, a novel \underline{Pa}ttern-\underline{C}lustered \underline{K}nowledge \underline{D}istillation approach to address the challenges.
First, we propose applying knowledge distillation for compressing memory access prediction models. 
Knowledge Distillation (KD)~\cite{hinton2015distilling} is a technique that transfers knowledge from a large model to a smaller model, enabling the smaller model to replicate the behavior and performance of the larger model.
This approach facilitates more efficient deployment and improves generalization, making it well-suited for MAP models.
In offline training, training time and model size are variable, but for online deployment, MAP models have the strict requirements of fast inference and efficient storage.
Therefore, we can train large teacher models and distill the knowledge to a compact student model offline while deploying the student model for online inference. 
Second, we propose to cluster the memory access patterns to train higher performance, pattern-specific teachers for each cluster. 
In this way, we are able to distill specialized knowledge from multiple teachers to a single student that works for all patterns within a specific trace. 
We demonstrate the effectiveness of our \ourwork\footnote{The code is available at: \url{https://github.com/neeleshg23/PaCKD}} through the widely used GAP Benchmark Suite~\cite{beamer2017gap}.

Our main contributions can be summarized as follows:
\begin{itemize}
    \item We present \ourwork, a novel knowledge distillation (KD) framework for compressing memory access prediction models while maintaining the performance.
    \item We propose to cluster the memory access sequences based on history memory access windows comprised of the past block addresses, past block address deltas, and past instruction pointers to increase the performance of DNN-based MAP models.
    \item We introduce an ensemble multi-label KD approach to transfer specialized knowledge from large pattern specific models to a single smaller and lightweight student model for the task of multi-label memory access prediction.
    \item We evaluate the effectiveness of \ourwork~on LSTM, MLP-Mixer, and ResNet using the GAP Benchmark. We achieve 552$\times$ model size compression in average with a 1.92\% F1-score drop, which is is 8.70\% higher compared with student models using standard KD and 8.88\% higher compared with student models trained without KD.
\end{itemize}

\begin{figure*}[ht]
  \centering
  \includegraphics[width=0.95\linewidth]{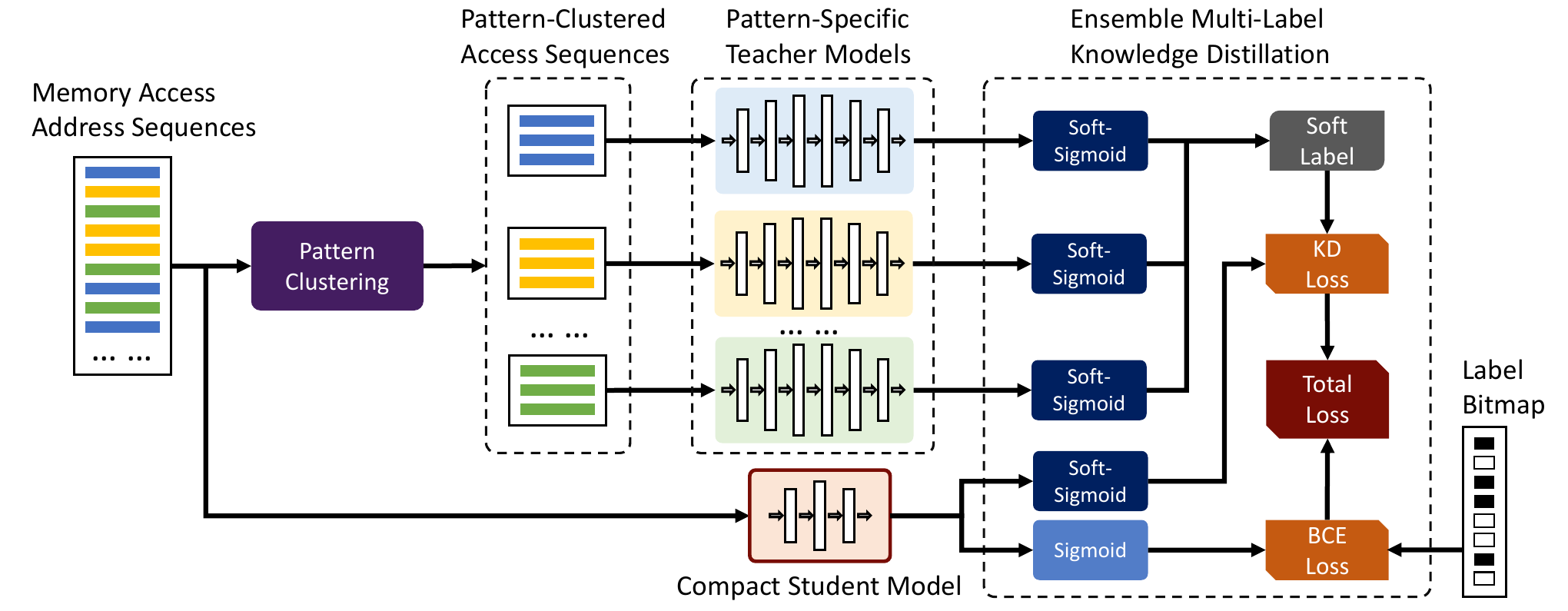}
  \caption{The Pattern-Clustered Knowledge Distillation (\ourwork) framework for compressing memory access prediction models.
  }
\label{fig:overview}
\end{figure*}

\section{Background}

\subsection{Memory Access Prediction using Machine Learning}

Memory access prediction is a critical task that involves correlating past memory accesses with the present, detecting patterns in memory access behavior, and accurately predicting future memory access addresses. This capability is crucial for enabling efficient data prefetching to minimize memory latency and improve overall system performance. Traditional prefetchers focus on exploiting spatial or temporal locality of memory references \cite{kumar1998exploiting, cucchiara2001temporal}. However, machine learning (ML) models have emerged as a promising approach for prefetching complex data access patterns. The problem definition of ML-based memory access prediction shown below.

\noindent{\textbf{Problem Definition.}} Let $X_t = \{x_1, x_2, ..., x_N\}$ be the sequence of $N$ history memory addresses at time $t$; let $Y_t=\{y_1, y_2, ..., y_k\}$ be a set of $k$ outputs that will be accessed in the future; an ML model can approximate $P(Y_t | X_t)$, the probability that the future addresses $Y_t$ will be accessed given the history events $X_t$.

Recent research has demonstrated the effectiveness of ML models in memory access prediction tasks \cite{zhang2022transfetch,zhang2022sharp}.
These existing approaches leverage fine-grained memory address input and attention-based models to achieve accurate multi-label memory access prediction.
By capturing intricate dependencies and patterns in memory access behavior, image models offer improvements in memory access prediction compared to traditional techniques \cite{zhang2022transfetch, srivastava2019predicting}.

While existing MAP models focus primarily on prediction performance, this work aims to make these models more hardware-friendly by reducing their size without sacrificing performance.

\subsection{Clustering Memory Access Sequences}
Clustering memory access sequences based on patterns is a valuable technique for identifying similar behaviors within these traces \cite{altun2006clustering,zhang2022c}. This approach groups together trace instructions with shared characteristics, further specializing models. Through finding patterns in each sequence, models are able to exploit patterns in order to improve performance.

\noindent{\textbf{Problem Definition.}} Given a set of $N$ memory access sequences \{{$\mathbf{x_1}$, $\mathbf{x_2}$, ..., $\mathbf{x_n}$\} and an integer $k$, a memory access sequence clustering algorithm is to partition the sequences to $k$ clusters \{$C_1$,$C_2$, ...,$C_k$\} such that sequences in the same cluster share similar patterns. 

We leverage K-Means~\cite{hartigan1979algorithm} algorithm for clustering a memory access sequence. K-means represents each cluster by a centroid which is the mean of the cluster members, then uses squared Euclidean distance to measure for cluster membership, as defined below:
\begin{equation}
    d_{sq} = \sum_{i=1}^{D} (x_i - y_i)^2
\end{equation}
where $x, y$ are points in the $D$-dimensional space. Number of clusters $k$ is determined by minimizing the Sum of Squared Errors (SSE), which is the sum of the squared error between each data point and its nearest centroid, as defined below:
\begin{equation}
    SSE = \sum_{i=1}^{n} \sum_{j=1}^{k} w_{i,j} ||x_i - c_j||^2
\end{equation}
where $c_j$ is the centroid of the $j^{th}$ cluster, $w_{i,j}=1$ if the data point $x_i$ is in cluster $j$, and $w_{i,j}=0$ otherwise \cite{bishop2006pattern}.

Building upon the existing clustering approaches, we cluster the memory access sequences based on different memory access features, aiming to create specialized clusters that capture distinct patterns and domain-knowledge within memory access traces to enable effective training.

\subsection{Knowledge Distillation}
Knowledge Distillation aims to transfer knowledge from a large, complex model (referred to as the teacher model) to a smaller, more lightweight model (referred to as the student model)~\cite{hinton2015distilling}. The student model is trained to mimic the behavior and predictions of the teacher model by learning the embedded information in the teacher's logits. This process has been shown to be effective in improving the performance of compact models in hopes of achieving comparable performance to their larger counterparts. The problem definition of using KD for model compression is as below.

\noindent{\textbf{Problem Definition.}} Given a trained large teacher network $f(x;\theta)$, where $x$ is the input to the network and $\theta$ is the parameters. The goal of knowledge distillation is to learn a new set of parameters $\theta'$ for
a shallower student network $f(x;\theta')$, such that the student network achieves similar performance to the teacher, with much lower computational cost.

The original knowledge distillation work primarily focuses on single-label classification. In this context, a custom loss function \(L\) is designed that considers both the hard labels from ground truth and soft labels from a teacher model. Importantly, the soft labels are obtained using a T-temperature softmax activation function.

The softmax activation function returns probabilities \(P(z_i, T)\) is defined as:

\begin{equation}
\label{eq:kd-4}
P(z_i, T) = \frac{\exp \left(\frac{z_i}{T}\right)}{\sum_{k=1}^{N} \exp \left(\frac{z_k}{T}\right)}
\end{equation}

Therefore, the complete loss \(L\) can be represented as:
\begin{equation}
\label{eq:kd-2}
L_{\text{soft}} = -\sum_{i=1}^{N} P(t_i, T) \log \left(P(s_i, T)\right)
\end{equation}
\begin{equation}
\label{eq:kd-3}
L_{\text{hard}} = -\sum_{i=1}^{N} P(c_i, 1) \log \left(P(s_i, 1)\right)
\end{equation}
\begin{equation}
\label{eq:kd-1}
L = \alpha L_{\text{soft}} + \beta L_{\text{hard}}
\end{equation}

where \( \alpha \) and \( \beta \) are hyperparameters, \( t_i \) are the logits from the teacher model, \( s_i \) are the logits from the student model, \( c_i \) is the ground truth for the \(i\)-th instance, \( N \) is the total number of instances, and \( T \) is the temperature parameter for softmax.

In this work, we extend the original KD approach by training pattern-specific teacher models and designing a soft-sigmoid activation function for multi-label knowledge distillation (Section~\ref{sec:mlkd}). By implementing the proposed approach in memory access prediction models, achieving efficient compression becomes possible by transferring the expertise of a large and intricate teacher model to a smaller and more streamlined student model. This approach allows for significant memory savings without compromising on performance and accuracy, as the student model can closely match the capabilities of the teacher model\cite{li2021deep}.

\section{Approach}

\subsection{Overview}

Existing DNN-based MAP models require large storage allocation, making them impractical for hardware data prefetchers. We propose \ourwork~to address such limitations.

\noindent{\textbf{Research Hypothesis.}} By leveraging featured clustering, pattern-specific teacher models, and multi-label ensemble knowledge distillation, we can effectively compress memory access prediction models, reducing their memory footprint while preserving their performance. 

Figure~\ref{fig:overview} shows an illustration of the proposed training framework,~\ourwork, for compressing memory access prediction models. Our approach involves three key steps. Firstly, we employ clustering by memory pattern-specific access trace data to partition memory access sequences into distinct clusters, effectively capturing underlying patterns. Secondly, we train large pattern-specific teacher models specializing in predicting memory access within each cluster, enhancing overall prediction performance. Finally, we employ knowledge distillation to train a lightweight student model, transferring novel insights from the pattern-specific teacher models. 
This process balances accuracy and computational cost, optimizing the training of compact memory access prediction models.

\subsection{Memory Access Sequence Clustering}

We use k-means for memory access sequence clustering, exploring various memory access features.

\begin{itemize}
    \item \textbf{Past Block Address}: The previous memory address accessed in the program's execution trace, representing a lookback window of one block for cache optimization.
    \item \textbf{Past Block Address Deltas}: 
    The numerical difference between consecutively accessed memory addresses in the trace plays a pivotal role in domain-specific hardware-level instructions since it reflects a lookback window essential for fine-tuning and enhancing cache access.
    \item \textbf{Past Instruction Pointer}: The memory address of the last executed instruction, providing a lookback for immediate hardware-level instruction tracking.
\end{itemize}

\label{sec:clustering}

\begin{figure*}[h]
  \centering
  \includegraphics[width=\linewidth]{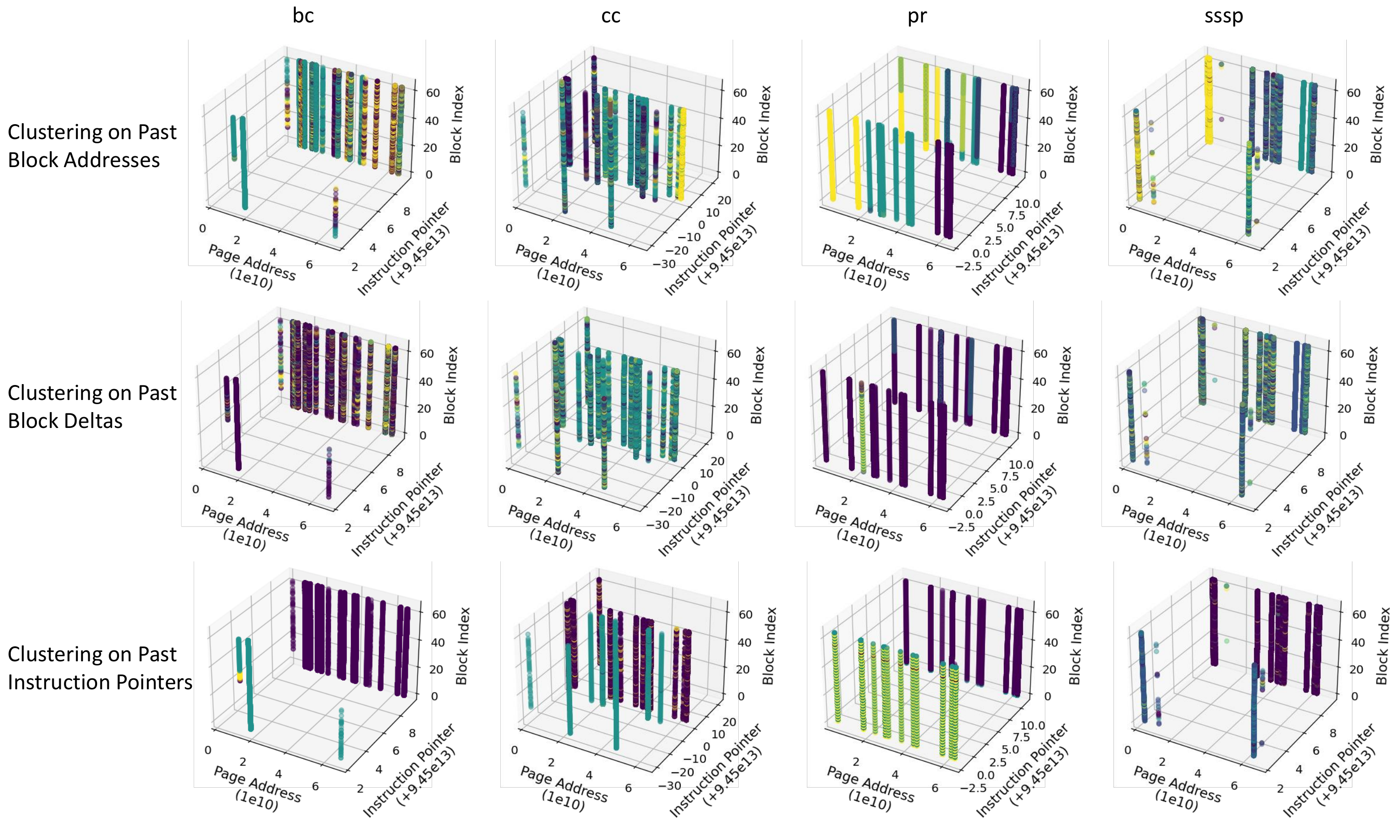}
  \caption{Clustering results for applications in GAP benchmark under various input features.}
\label{fig:cluster}
\end{figure*}

\subsection{Training Pattern-Specific Teacher Models}
After clustering memory access sequences using memory trace features that embed application-specific information, we train large pattern-specific teacher models for each cluster.

\noindent{\textbf{Input.}} We use segmented memory access addresses as model input, following TransFetch~\cite{zhang2022transfetch}. This approach splits an $m$-bit block address into $p$ segments, converting one block address into a vector. For a history of $n$ lookback block addresses, the input would be a $p\times n$ matrix.

\noindent{\textbf{Output.}} For model training labels, we use a delta bitmap. In this bitmap, each location represents the difference between a future memory access address and the current address. A value of 1 is assigned to a location if the delta falls within a future window, otherwise, it is set to 0. By setting multiple 1s in the bitmap, we approach the machine learning problem as multi-label classification, enabling multiple predictions.

\noindent{\textbf{Training.}}
We use binary cross-entropy loss to train the multi-label classification model, as defined below:
\begin{equation}
\mathcal{L}_{\text {BCE}}=-\frac{1}{N} \sum_{i=1}^{N} y_{i} \log \left(p\left(y_{i}\right)\right)+\left(1-y_{i}\right) \log \left(1-p\left(y_{i}\right)\right)
\end{equation}
where $y_i$ is the label and $p(y_i)$ is the predicted probability for sample $i$ being 1, $N$ is the number of classes. 

\subsection{Ensemble Multi-Label Knowledge Distillation}

We align the input and output formats of the student models with those of the teacher models and train them through a novel multi-label KD approach, leveraging an ensemble training scheme based on multiple teachers.
\subsubsection{Multi-Label Knowledge Distillation}
\label{sec:mlkd}
Inspired by~\cite{hinton2015distilling}, we design a soft-sigmoid function in Equation~\ref{eq:sigmoid} with temperature $T$ to soften the probability distribution over classes in multi-label classification outputs. The overall loss function is the combination of the BCE loss and the soft KD loss (Equation~\ref{eq:kd_loss}) acquired from the soft-sigmoid, as shown in Equation~\ref{eq:total_loss}

\begin{equation}
\begin{aligned}
\label{eq:sigmoid}
z_i=p(y_i)_{t=T}=\sigma\left(\frac{y_i}{T}\right)=\frac{1}{1+e^{-y_i/T}}
\end{aligned}
\end{equation}

\begin{equation}
\label{eq:kd_loss}
\mathcal{L}_{\mathrm{KD}}=\sum_{k=1}^q \text{KL}\left(\left[z_i^{\mathcal{T}}, 1-z_i^{\mathcal{T}}\right] \|\left[z_i^{\mathcal{S}}, 1-z_i^{\mathcal{S}}\right]\right)
\end{equation}
where $KL(\cdot)$ is the Kullback-Leibler divergence~\cite{joyce2011kullback}, $\lambda$ is a hyper-parameter tuning the weights of the two losses.

\begin{equation}
\label{eq:total_loss}
\mathcal{L}=\lambda \mathcal{L}_{\text {KD}} + (1-\lambda)\mathcal{L}_{\text {BCE}}
\end{equation}

\subsubsection{Ensemble Training for a Single Student} 

In order to train a single student model capable of accurately predicting the entire memory access sequence, we employ knowledge distillation from each teacher during every training epoch. The distillation process utilizes the same clustered data that was used to train the teachers. The weighting of the teacher's logits in the overall loss of the student model is determined by $\lambda$, which is carefully tuned to assign greater importance to teachers with more precise predictions.

\section{Evaluation}

\subsection{Experimental Setup}

\subsubsection{Models}

We evaluate the performance of \ourwork~on three state-of-the-art DNN models chosen based on their diverse underlying structures.
This evaluation allows us to assess the how general \ourwork~works across different model architectures.
We underscore the usage of any image, vision, or multi-label classification model in critical system tasks, including, but not limited to, memory access prediction. 

\begin{itemize}
    \item \textbf{Long Short-Term Memory (L)}: LSTM-based models have been successful in both single-label and multi-label prediction tasks and are widely adopted due to high applicability. LSTM models capture long-range features along with temporal patterns \cite{hochreiter1997long, tatsunami2022sequencer}.
    \item \textbf{Multi-Layer Perceptron Mixer (M)}: Without convolutions or attention, an all MLP-based model, made of two types of layers: channel-mixing and token-mixing layers, allows effective model learning that mixes both spatial and temporal information \cite{tolstikhin2021mlp}.
    \item \textbf{Residual Networks (R)}: ResNets are Convolution Neural Networks (CNN) which rely on a residual learning framework to transfer knowledge to deeper layers over time, improving with increases in depth and size \cite{he2016residual}.
\end{itemize}

Table \ref{tab:model_data} presents specific student and teacher model configurations, including paramters, dimensions, number of layers, and critical path, where $N$ represents the number of time steps and $L$ represents the number of layers in the model.

The average teacher model size is 5.406M parameters and the average student model size is 10.487K parameters.
The compression rate for the two student models are 445$\times$, 538$\times$, and 584$\times$, and the average compression rate is 552$\times$.

\begin{table}[htb!]
  \caption{Model Configurations}
  \begin{tabular}{|c|c|c|c|c|c|}
    \hline
    \textbf{Model} & \textbf{Parameters} & \textbf{Dimensions} & \textbf{Layers} & \textbf{Critical Path}  \\
    \hline
    Tch-L & 5.302M & 128 & 40& $\mathcal{O}(NL)$\\
    Tch-M & 5.484M & 512 & 20&$\mathcal{O}(L)$\\
    Tch-R & 5.423M & 30 & 50&$\mathcal{O}(L)$\\
    Stu-L & 11.904K & 16 & 1&$\mathcal{O}(NL)$\\
    Stu-M & 10.206K & 18 & 8&$\mathcal{O}(L)$\\
    Stu-R & 9.324K & 4 & 11&$\mathcal{O}(L)$\\
    \hline 
    \multicolumn{5}{l}{\textit{Note: L means LSTM, M means MLPMixer, R means ResNet}}
  \end{tabular}
  \label{tab:model_data}
\end{table}

\subsubsection{Benchmarks}

To evaluate \ourwork~and the baselines, we use four real-world graph analytic benchmarks from the GAP Benchmark Suite~\cite{beamer2017gap}.
For each application, we capture memory requests from the last level cache over a duration of 11 million instructions. The initial one million instructions are skipped, while the subsequent 8 million instructions are used for training while the remaining 2 million instructions are used for evaluation.
Table \ref{tab:benchmarks} shows the application-specific memory trace sequence level statistics, including the unique number of instruction pointers, block addresses, and block address deltas. 

\begin{table}[htb!]
  \centering
  \caption{Benchmark Application Memory Trace Statistics}
  \begin{tabular}{|c|c|c|c|c|}
    \hline
    \textbf{Trace} & \textbf{Description} & \textbf{\# IPs} & \textbf{\# Addr} & \textbf{\# Delta} \\
    \hline
    bc & Betweenness Centrality & 78 & 144.2K & 52.6K \\
    cc & Connected Components & 45 & 95.5K & 80.1K \\
    pr & PageRank & 38 & 186.4K & 2.6K \\
    sssp & Single-Source Shortest Path & 102 & 86.7K & 66.8K \\
    \hline
  \end{tabular}
  \label{tab:benchmarks}
\end{table}
\subsubsection{Metrics}

We use F1-score to evaluate the memory access prediction performance, F1-score is the weighted average of precision and recall\cite{powers2008evaluation}.

\begin{equation}
\label{eq:precision}
\text{Precision} = \frac{\text{True Positives}}{\text{True Positives} + \text{False Positives}}
\end{equation}

\begin{equation}
\label{eq:recall}
\text{Recall} = \frac{\text{True Positives}}{\text{True Positives} + \text{False Negatives}}
\end{equation}

\begin{equation}
\label{eq:f1}
F_1 = 2 \times \frac{\text{Precision} \times \text{Recall}}{\text{Precision} + \text{Recall}}
\end{equation}

\begin{table}[ht!]
\centering
\caption{F1-Score of the teacher models and student models trained without Knowledge Distillation and Clustering}
  \begin{tabular}{|c|c|c|c|c|c|c|c|c|c|c|c|c|c|c|c|c|}
    \hline
    \textbf{Models}&{\textbf{bc}}&{\textbf{cc}}&{\textbf{pr}}&{\textbf{sssp}}\\ 
    \hline
    Tch-L & 0.3580 & 0.2167 & 0.7719 & 0.2411 \\
    \hline 
    Tch-M & 0.3717  & 0.3002 & 0.9754 & 0.4134 \\
    \hline
    Tch-R & 0.3817  & 0.3095 & 0.8932 & 0.3178 \\
    \hline
    Stu-L & 0.3580  & 0.2213 & 0.7723 & 0.2165 \\
    \hline
    Stu-M & 0.3579  & 0.2855 & 0.8803  & 0.2338 \\
    \hline
    Stu-R & 0.3575 & 0.2225 & 0.8629 & 0.2334 \\
    \hline
\end{tabular}
\label{tab:baseline}
\end{table}

\begin{table*}[htbp]
\centering
\caption{F1-score of ensemble teacher models, trained on traces clustered using different memory access features}
\begin{tabular}{|c|c||c|c|c||c|c|c||c|c|c||c|c|c|}
\hline
    \textbf{Memory Access}&&\multicolumn{3}{c||}{\textbf{bc}}&\multicolumn{3}{c||}{\textbf{cc}}&\multicolumn{3}{c||}{\textbf{pr}}&\multicolumn{3}{c|}{\textbf{sssp}}\\ 
    \cline{3-14}
    \textbf{Features}&\textbf{$K$}&\textbf{Stu-L}&\textbf{Stu-M}&\textbf{Stu-R}&\textbf{Stu-L}&\textbf{Stu-M}&\textbf{Stu-R}&\textbf{Stu-L}&\textbf{Stu-M}&\textbf{Stu-R}&\textbf{Stu-L}&\textbf{Stu-M}&\textbf{Stu-R}\\
    \hline
    & 2 & 0.2226 & 0.3624 & 0.3573 & 0.2167 & 0.3080 & 0.3397 & 0.7717 & \textbf{0.9897} & 0.8006 & 0.2316 & 0.4151 & 0.3223 \\
    Past Block Address & 3 & 0.3575 & 0.3640 & 0.3570 & 0.2156 & \textbf{0.3316} & 0.2787 & 0.7687 & 0.9658 & 0.8090 & 0.2364 & 0.3834 & \textbf{0.3876} \\
    & 4 & 0.3578 & 0.3627 & 0.3589 & 0.2096 & 0.3247 & \textbf{0.3655} & 0.7689 & 0.9779 & 0.8251 & 0.2317 & 0.3741 & 0.2890 \\

    \hline
    & 2 & 0.2836 & 0.2917 & 0.2855 & 0.1267 & 0.3214 & 0.2983 & 0.7709 & 0.8472 & 0.8306 & 0.1851 & 0.2882 & 0.3319 \\
    Past Address Delta & 3 & 0.2603 & 0.2748 & 0.2792 & 0.0585 & 0.1300 & 0.1040 & 0.8339 & 0.9693 & 0.9477 & 0.0883 & 0.2249 & 0.1717 \\
    & 4 & 0.2422 & 0.2456 & 0.2487 & 0.0661 & 0.1285 & 0.0971 & 0.8287 & 0.9651 & 0.9466 & 0.0847 & 0.1744 & 0.1778\\\cline{2-14}

    \hline
    & 2 & \textbf{0.3685} & \textbf{0.3926} & \textbf{0.3824} & 0.2169 & 0.2981 & 0.2894 & 0.7035 & 0.8706 & 0.8687 & 0.2422 & 0.4071 & 0.2959 \\
    Past Instruction Pointer & 3 & 0.3621 & 0.3687 & 0.3696 & \textbf{0.2186} & 0.3040 & 0.2803 & \textbf{0.8375} & 0.9876 & \textbf{0.9705} & \textbf{0.2506} & \textbf{0.4223} & 0.2880 \\
    & 4 & 0.3621 & 0.3706 & 0.3689 & 0.2117 & 0.2977 & 0.2967 & 0.8359 & 0.9874 & 0.9641 & 0.2505 & 0.4149 & 0.2849 \\
    \hline
\end{tabular}
\label{tab:teacher}
\end{table*}

\begin{table*}[htbp]
\centering
\caption{F1-score of student models, trained on traces clustered using different memory access features}
\begin{tabular}{|c|c||c|c|c||c|c|c||c|c|c||c|c|c|}
\hline
    \textbf{Memory Access}&&\multicolumn{3}{c||}{\textbf{bc}}&\multicolumn{3}{|c||}{\textbf{cc}}&\multicolumn{3}{c||}{\textbf{pr}}&\multicolumn{3}{c|}{\textbf{sssp}}\\ 
    \cline{3-14}
    \textbf{Features}&\textbf{$K$}&\textbf{Stu-L}&\textbf{Stu-M}&\textbf{Stu-R}&\textbf{Stu-L}&\textbf{Stu-M}&\textbf{Stu-R}&\textbf{Stu-L}&\textbf{Stu-M}&\textbf{Stu-R}&\textbf{Stu-L}&\textbf{Stu-M}&\textbf{Stu-R}\\
    \hline 
    
    Standard KD & 1 & \textbf{0.3580} & 0.3579 & 0.3575 & 0.2234 & 0.2845 & 0.2216 & 0.7723 & 0.8774 & 0.8731 & 0.2168 & 0.2340 & 0.2336 \\\cline{2-14}
    \hline
    & 2 & 0.3580 & 0.3579 & \textbf{0.3587} & 0.2216 & 0.2866 & 0.3076 & 0.7724 & 0.9305 & 0.9525 & 0.2426 & 0.2683 & 0.2710 \\
    Past Block Address & 3 & 0.3580 & 0.3578 & 0.3577 & 0.2212 & 0.2898 & 0.2939 & 0.7725 & 0.9268 & 0.9518 & \textbf{0.2429} & 0.2422 & \textbf{0.3323}\\
    & 4 & 0.3580 & 0.3579 & 0.3580 & 0.2215 & 0.2807 & 0.2209 & 0.7724 & 0.8595 & 0.8339 & 0.2164 & 0.2316 & 0.2269 \\\cline{2-14}
    \hline
    & 2 & 0.3580 & 0.3580 & 0.3579 & \textbf{0.2216} & 0.2931 & \textbf{0.3196} & 0.7724 & 0.8860 & 0.9222 & 0.2429 & 0.2652 & 0.2938 \\
    Past Address Delta & 3 & 0.3580 & 0.3578 & 0.3580 & 0.2216 & \textbf{0.2957} & 0.3087 & 0.7724 & 0.7742 & 0.8147 & 0.2429 & \textbf{0.2705} & 0.2489 \\
    & 4 & 0.3580 & 0.3580 & 0.3579 & 0.2214 & 0.2876 & 0.2342 & 0.7719 & 0.7735 & 0.7851 & 0.2168 & 0.2347 & 0.2336\\\cline{2-14}
    \hline
    & 2 & 0.3580 & \textbf{0.3582} & 0.3577 & 0.2216 & 0.2913 & 0.3196 & \textbf{0.7727} & \textbf{0.9519} & \textbf{0.9639} & 0.2426 & 0.2352 & 0.2466 \\
    Past Instruction Pointer & 3 & 0.3580 & 0.3579 & 0.3580 & 0.2216 & 0.2957 & 0.3114 & 0.7723 & 0.7742 & 0.8138 & 0.2422 & 0.2343 & 0.2418 \\
    & 4 & 0.3580 & 0.3580 & 0.3580 & 0.2214 & 0.2876 & 0.2343 & 0.7719 & 0.7734 & 0.7854 & 0.2168 & 0.2316 & 0.2326 \\\cline{2-14}
    \hline
\end{tabular}
\label{tab:student}
\end{table*}

\subsection{Baseline Frameworks} 

We demonstrate the effectiveness of the proposed approach by comparing \ourwork~with the following baseline frameworks:

\begin{itemize}
    \item Student Only: Trained without knowledge distillation, this serves as a measure of the baseline performance.
    \item Teacher Only: Larger models trained without clustering, enabling us to gauge the influence of our pattern-driven clustering approach.
    \item Standard KD Student: Smaller models trained with a single teacher, using a $\lambda$ value of $0.5$, allow us to evaluate the benefit of ensemble knowledge distillation.
\end{itemize}

\subsection{Clustering and Training}
Based on the features of each memory access trace, we split the application into $K$ distinct clusters.
Figure~\ref{fig:cluster} shows the clustering results for applications in GAP benchmark under various input features when K=3.
The clusters are visualized using dimensions of Page Address, Instruction Pointer, and Block Index with respective scale and bias shown with the axis labels.
Various input features significantly influence the clustering results, influencing the training of pattern-specific teacher. 
For the prediction models, we set the lookback window as 10. We collect future deltas within a page in a window of 128 future memory accesses. We predict deltas in a range of $\pm 128$.
We generate train and evaluation data for each cluster to be given to teacher models, as well as evaluation data over the entire cluster to validate the student model. 

\subsection{Result Analysis}

\subsubsection{Effectiveness of Clustering for Teacher Training}
Clustering significantly enhanced our teacher models. As shown in Table~\ref{tab:baseline}, the average F1-score started at 0.4626 across all models and applications. After clustering, the score rose to 0.4931 as shown in Table~\ref{tab:teacher}, marking a 6.61\% improvement, showing the effectiveness of clustering in our approach.

\subsubsection{Effectiveness of Standard Knowledge Distillation}
Table \ref{tab:student} reveals that student models trained with standard KD exhibit a marginal increase of 0.16\% in F1-score compared to the baseline student models without distillation (Table \ref{tab:baseline}).

\subsubsection{Effectiveness of Ensemble Knowledge Distillation}
Table \ref{tab:student} shows that student models trained using ensemble KD achieve an average F1-score of 0.4538, surpassing the performance of student models trained with standard KD by 8.70\%.

\subsubsection{Effectiveness of \ourwork}
\ourwork~demonstrates its effectiveness through achieving an average F1-score increase of 8.88\% compared to baseline student models trained without KD. Additionally, \ourwork~achieves a compression rate of 522$\times$ with only a 1.92\% drop in F1-score compared to the baseline teacher models without clustering.

\section{Conclusion}

In this paper, we introduced \ourwork, a novel approach for compressing memory access prediction models.
Our method utilizes pattern-clustered knowledge distillation, leveraging clustering for improved performance in large pattern-specific teacher models.
By distilling specialized knowledge from multiple teachers to a lightweight student model, we achieved an impressive compression ratio of 552$\times$ with only a 1.92\% drop in F1-score.
This result outperforms standard knowledge distillation by 8.70\% and direct student training by 8.88\%. 

To improve the practicality of ML-based models for data prefetching, we are also exploring novel modeling of memory accesses, such as graph representation of memory accesses and graph neural networks for memory access prediction. We are also exploring efficient parallel implementation and look-up table approximation for neural networks for real-time ML-based prefetching. In future work, we hope to extend our approach to heterogeneous memory systems and heterogeneous clusters. 

\section*{Acknowledgment}
This work was supported by Army Resarch Lab (ARL) under award number W911NF2220159 and National Science Foundation (NSF) under grants under award number CCF-1919289.

\bibliographystyle{IEEEtran}
\bibliography{my_ref}

\begin{thebibliography}{10}
\providecommand{\url}[1]{#1}
\csname url@samestyle\endcsname
\providecommand{\newblock}{\relax}
\providecommand{\bibinfo}[2]{#2}
\providecommand{\BIBentrySTDinterwordspacing}{\spaceskip=0pt\relax}
\providecommand{\BIBentryALTinterwordstretchfactor}{4}
\providecommand{\BIBentryALTinterwordspacing}{\spaceskip=\fontdimen2\font plus
\BIBentryALTinterwordstretchfactor\fontdimen3\font minus
  \fontdimen4\font\relax}
\providecommand{\BIBforeignlanguage}[2]{{%
\expandafter\ifx\csname l@#1\endcsname\relax
\typeout{** WARNING: IEEEtran.bst: No hyphenation pattern has been}%
\typeout{** loaded for the language `#1'. Using the pattern for}%
\typeout{** the default language instead.}%
\else
\language=\csname l@#1\endcsname
\fi
#2}}
\providecommand{\BIBdecl}{\relax}
\BIBdecl

\bibitem{wulf1995hitting}
W.~A. Wulf and S.~A. McKee, ``Hitting the memory wall: Implications of the
  obvious,'' \emph{ACM SIGARCH computer architecture news}, vol.~23, no.~1, pp.
  20--24, 1995.

\bibitem{carvalho2002gap}
C.~Carvalho, ``The gap between processor and memory speeds,'' in \emph{Proc. of
  IEEE International Conference on Control and Automation}, 2002.

\bibitem{baer1991cache}
J.-L. Baer and T.-F. Chen, ``An effective on-chip preloading scheme to reduce
  data access penalty,'' in \emph{Supercomputing '91:Proceedings of the 1991
  ACM/IEEE Conference on Supercomputing}, 1991, pp. 176--186.

\bibitem{wiel1997prefetch}
S.~Vander~Wiel and D.~Lilja, ``When caches aren't enough: data prefetching
  techniques,'' \emph{Computer}, vol.~30, no.~7, pp. 23--30, 1997.

\bibitem{choi2021survey}
\BIBentryALTinterwordspacing
H.~Choi and S.~Park, ``A survey of machine learning-based system performance
  optimization techniques,'' \emph{Applied Sciences}, vol.~11, no.~7, 2021.
  [Online]. Available: \url{https://www.mdpi.com/2076-3417/11/7/3235}
\BIBentrySTDinterwordspacing

\bibitem{hashemi2018mempatterns}
\BIBentryALTinterwordspacing
M.~Hashemi, K.~Swersky, J.~A. Smith, G.~Ayers, H.~Litz, J.~Chang, C.~Kozyrakis,
  and P.~Ranganathan, ``Learning memory access patterns,'' \emph{CoRR}, vol.
  abs/1803.02329, 2018. [Online]. Available:
  \url{http://arxiv.org/abs/1803.02329}
\BIBentrySTDinterwordspacing

\bibitem{srivastava2019predicting}
A.~Srivastava, A.~Lazaris, B.~Brooks, R.~Kannan, and V.~K. Prasanna,
  ``Predicting memory accesses: the road to compact ml-driven prefetcher,'' in
  \emph{Proceedings of the International Symposium on Memory Systems}, 2019,
  pp. 461--470.

\bibitem{zeng2017lstm}
\BIBentryALTinterwordspacing
Y.~Zeng, ``Long short term based memory hardware prefetcher,'' 2017. [Online].
  Available: \url{http://preserve.lehigh.edu/etd/2901}
\BIBentrySTDinterwordspacing

\bibitem{hooker2019compressed}
S.~Hooker, A.~Courville, G.~Clark, Y.~Dauphin, and A.~Frome, ``What do
  compressed deep neural networks forget?'' \emph{arXiv preprint
  arXiv:1911.05248}, 2019.

\bibitem{lakhotia2020gpop}
K.~Lakhotia, R.~Kannan, S.~Pati, and V.~Prasanna, ``Gpop: A scalable cache-and
  memory-efficient framework for graph processing over parts,'' \emph{ACM
  Transactions on Parallel Computing}, vol.~7, no.~1, pp. 1--24, 2020.

\bibitem{xstream}
A.~Roy, I.~Mihailovic, and W.~Zwaenepoel, ``X-stream: Edge-centric graph
  processing using streaming partitions,'' in \emph{Proceedings of the
  Twenty-Fourth ACM Symposium on Operating Systems Principles}.\hskip 1em plus
  0.5em minus 0.4em\relax ACM, 2013, pp. 472--488.

\bibitem{pcpm}
K.~Lakhotia, R.~Kannan, and V.~Prasanna, ``Accelerating pagerank using
  partition-centric processing,'' in \emph{2018 USENIX Annual Technical
  Conference (USENIX ATC 18)}.\hskip 1em plus 0.5em minus 0.4em\relax USENIX
  Association, 2018.

\bibitem{french1999catastrophic}
\BIBentryALTinterwordspacing
R.~M. French, ``Catastrophic forgetting in connectionist networks,''
  \emph{Trends in Cognitive Sciences}, vol.~3, no.~4, pp. 128--135, 1999.
  [Online]. Available:
  \url{https://www.sciencedirect.com/science/article/pii/S1364661399012942}
\BIBentrySTDinterwordspacing

\bibitem{kirkpatrick2017overcoming}
J.~Kirkpatrick, R.~Pascanu, N.~Rabinowitz, J.~Veness, G.~Desjardins, A.~Rusu,
  K.~Milan, J.~Quan, T.~Ramalho, A.~Grabska-Barwinska, D.~Hassabis, C.~Clopath,
  D.~Kumaran, and R.~Hadsell, ``Overcoming catastrophic forgetting in neural
  networks,'' \emph{Proceedings of the National Academy of Sciences}, vol. 114,
  12 2016.

\bibitem{hinton2015distilling}
G.~Hinton, O.~Vinyals, and J.~Dean, ``Distilling the knowledge in a neural
  network,'' \emph{arXiv preprint arXiv:1503.02531}, 2015.

\bibitem{beamer2017gap}
S.~Beamer, K.~Asanović, and D.~Patterson, ``The gap benchmark suite,'' 2017.

\bibitem{kumar1998exploiting}
S.~Kumar and C.~Wilkerson, ``Exploiting spatial locality in data caches using
  spatial footprints,'' in \emph{Proceedings. 25th Annual International
  Symposium on Computer Architecture (Cat. No. 98CB36235)}.\hskip 1em plus
  0.5em minus 0.4em\relax IEEE, 1998, pp. 357--368.

\bibitem{cucchiara2001temporal}
R.~Cucchiara, M.~Piccardi, and A.~Prati, ``Temporal analysis of cache
  prefetching strategies for multimedia applications,'' in \emph{Conference
  Proceedings of the 2001 IEEE International Performance, Computing, and
  Communications Conference (Cat. No.01CH37210)}, 2001, pp. 311--318.

\bibitem{zhang2022transfetch}
\BIBentryALTinterwordspacing
P.~Zhang, A.~Srivastava, A.~V. Nori, R.~Kannan, and V.~K. Prasanna,
  ``Fine-grained address segmentation for attention-based variable-degree
  prefetching,'' in \emph{Proceedings of the 19th ACM International Conference
  on Computing Frontiers}, ser. CF '22.\hskip 1em plus 0.5em minus 0.4em\relax
  New York, NY, USA: Association for Computing Machinery, 2022, p. 103–112.
  [Online]. Available:
  \url{https://doi-org.libproxy2.usc.edu/10.1145/3528416.3530236}
\BIBentrySTDinterwordspacing

\bibitem{zhang2022sharp}
P.~Zhang, R.~Kannan, X.~Tong, A.~V. Nori, and V.~K. Prasanna, ``Sharp: Software
  hint-assisted memory access prediction for graph analytics,'' in \emph{2022
  IEEE High Performance Extreme Computing Conference (HPEC)}.\hskip 1em plus
  0.5em minus 0.4em\relax IEEE, 2022, pp. 1--8.

\bibitem{altun2006clustering}
O.~Altun, N.~Dursunoglu, and M.~F. Amasyali, ``Clustering application
  benchmark,'' in \emph{2006 IEEE International Symposium on Workload
  Characterization}, 2006, pp. 178--181.

\bibitem{zhang2022c}
P.~Zhang, A.~Srivastava, T.-Y. Wang, C.~A. De~Rose, R.~Kannan, and V.~K.
  Prasanna, ``C-memmap: clustering-driven compact, adaptable, and generalizable
  meta-lstm models for memory access prediction,'' \emph{International Journal
  of Data Science and Analytics}, vol.~13, no.~1, pp. 3--16, 2022.

\bibitem{hartigan1979algorithm}
J.~A. Hartigan and M.~A. Wong, ``Algorithm as 136: A k-means clustering
  algorithm,'' \emph{Journal of the royal statistical society. series c
  (applied statistics)}, vol.~28, no.~1, pp. 100--108, 1979.

\bibitem{bishop2006pattern}
C.~M. Bishop, \emph{Pattern Recognition and Machine Learning (Information
  Science and Statistics)}.\hskip 1em plus 0.5em minus 0.4em\relax Berlin,
  Heidelberg: Springer-Verlag, 2006.

\bibitem{li2021deep}
\BIBentryALTinterwordspacing
X.~Li, H.~Xiong, Z.~Chen, J.~Huan, C.-Z. Xu, and D.~Dou, ``“in-network
  ensemble”: Deep ensemble learning with diversified knowledge
  distillation,'' vol.~12, no.~5, dec 2021. [Online]. Available:
  \url{https://doi.org/10.1145/3473464}
\BIBentrySTDinterwordspacing

\bibitem{joyce2011kullback}
J.~M. Joyce, ``Kullback-leibler divergence,'' in \emph{International
  encyclopedia of statistical science}.\hskip 1em plus 0.5em minus 0.4em\relax
  Springer, 2011, pp. 720--722.

\bibitem{hochreiter1997long}
S.~Hochreiter and J.~Schmidhuber, ``Long short-term memory,'' \emph{Neural
  Computation}, vol.~9, no.~8, pp. 1735--1780, 1997.

\bibitem{tatsunami2022sequencer}
Y.~Tatsunami and M.~Taki, ``Sequencer: Deep lstm for image classification,''
  \emph{Advances in Neural Information Processing Systems}, 2022.

\bibitem{tolstikhin2021mlp}
I.~O. Tolstikhin, N.~Houlsby, A.~Kolesnikov, L.~Beyer, X.~Zhai, T.~Unterthiner,
  J.~Yung, A.~Steiner, D.~Keysers, J.~Uszkoreit, M.~Lucic, and A.~Dosovitskiy,
  ``Mlp-mixer: An all-mlp architecture for vision,'' in \emph{Advances in
  Neural Information Processing Systems}, M.~Ranzato, A.~Beygelzimer,
  Y.~Dauphin, P.~Liang, and J.~W. Vaughan, Eds., vol.~34.\hskip 1em plus 0.5em
  minus 0.4em\relax Curran Associates, Inc., 2021, pp. 24\,261--24\,272.

\bibitem{he2016residual}
K.~He, X.~Zhang, S.~Ren, and J.~Sun, ``Deep residual learning for image
  recognition,'' in \emph{2016 IEEE Conference on Computer Vision and Pattern
  Recognition (CVPR)}, 2016, pp. 770--778.

\bibitem{powers2008evaluation}
D.~Powers, ``Evaluation: From precision, recall and f-factor to roc,
  informedness, markedness \& correlation,'' \emph{Mach. Learn. Technol.},
  vol.~2, 01 2008.

\end{thebibliography}

\end{document}